\documentclass[11pt,a4paper]{article}
\usepackage{times,latexsym}
\usepackage{url}
\usepackage[T1]{fontenc}
\usepackage{array}
\usepackage{amsmath}
\usepackage{caption}
\usepackage{subcaption}
\usepackage{booktabs}
\usepackage{amssymb}
\usepackage{multirow}
\usepackage{graphicx}
\usepackage{dsfont}
\usepackage{balance}
\usepackage{enumitem}
\usepackage{xcolor,colortbl}
\usepackage{makecell}
\usepackage{soul}
\usepackage{blindtext}
\usepackage[linesnumbered,ruled]{algorithm2e}
\usepackage{pifont}

\newcommand{\cmark}{\ding{51}}%
\newcommand{\xmark}{\ding{55}}%

\DeclareMathOperator*{\argmax}{\arg\!\max}
\usepackage[acceptedWithA]{tacl2018v2}
\usepackage[]{tacl2018v2}

\usepackage{xspace,mfirstuc,tabulary}

\newif\iftaclinstructions
\taclinstructionsfalse 
\iftaclinstructions
\renewcommand{\confidential}{}
\renewcommand{\anonsubtext}{(No author info supplied here, for consistency with
TACL-submission anonymization requirements)}
\newcommand{\instr}
\fi

\iftaclpubformat 

\else

\fi

\title{Syntax-guided Controlled Generation of Paraphrases}

\author{Ashutosh Kumar$^1$   \quad Kabir Ahuja$^{2*}$ \quad Raghuram Vadapalli$^3$\thanks{\, This research was conducted during the author's internship at Indian Institute of Science.}  \quad \textbf{Partha Talukdar}$^1$\\
	$^1$ Indian Institute of Science, Bangalore\\
	$^2$ Microsoft Research, Bangalore\\
	$^3$ Google, London\\
	{\tt \small ashutosh@iisc.ac.in, kabirahuja2431@gmail.com} \\
    {\tt \small raghuram.4350@gmail.com, ppt@iisc.ac.in} \\
}
\date{}

\begin{document}
\maketitle
\newcommand{\refalg}[1]{Algorithm \ref{#1}}
\newcommand{\refeqn}[1]{Equation \ref{#1}}
\newcommand{\reffig}[1]{Figure \ref{#1}}
\newcommand{\reftbl}[1]{Table \ref{#1}}
\newcommand{\refsec}[1]{Section \ref{#1}}

\newcommand{\methodmain}{\textsc{Sgcp}}
\newcommand{\scpn}{\textsc{Scpn}}
\newcommand{\cgen}{\textsc{Cgen}}
\newcommand{\methodfull}{\textbf{S}yntax \textbf{G}uided \textbf{C}ontrolled \textbf{P}araphraser}

\newcommand{\reminder}[1]{\textcolor{red}{[[ #1 ]]}\typeout{#1}}
\newcommand{\reminderR}[1]{\textcolor{gray}{[[ #1 ]]}\typeout{#1}}

\newcommand{\remove}[1]{\textcolor{red}{[[REMOVE: #1 ]]}\typeout{#1}}
\newcommand{\add}[1]{\textcolor{red}{#1}\typeout{#1}}

\newcommand{\m}[1]{\mathcal{#1}}

\newtheorem{theorem}{Theorem}[section]
\newtheorem{claim}[theorem]{Claim}

\newcommand{\tensor}{\mathcal{X}}
\newcommand{\Real}{\mathbb{R}}
\newcommand{\tuples}{\mathbb{T}}

\newcommand\norm[1]{\left\lVert#1\right\rVert}

\newcommand{\note}[1]{\textcolor{blue}{#1}}

\newcommand*{\Scale}[2][4]{\scalebox{#1}{$#2$}}%
\newcommand*{\Resize}[2]{\resizebox{#1}{!}{$#2$}}%

\def\mat#1{\mbox{\bf #1}}
\newcolumntype{P}[1]{>{\centering\arraybackslash}p{#1}} 

\begin{abstract}

    Given a sentence (e.g., \textit{"I like mangoes"}) and a constraint (e.g., sentiment flip), the goal of controlled text generation is to produce a sentence that adapts the input sentence to meet the requirements of the constraint (e.g., \textit{"I hate mangoes"}).
Going beyond such simple constraints, recent works have started exploring the incorporation of complex syntactic-guidance as constraints in the task of controlled paraphrase generation. In these methods, syntactic-guidance is sourced from a separate exemplar sentence. However, these prior works have only utilized limited syntactic information available in the parse tree of the exemplar sentence. We address this limitation in the paper and propose \methodfull{} (\methodmain{}), an end-to-end framework for syntactic paraphrase generation. We find that \methodmain{} can generate syntax-conforming sentences while not compromising on relevance. We perform extensive automated and human evaluations over multiple real-world English language datasets to demonstrate the efficacy of \methodmain{} over state-of-the-art baselines. To drive future research, we have made \methodmain{}'s source code available\footnote{\href{https://github.com/malllabiisc/SGCP}{https://github.com/malllabiisc/SGCP}}.
\end{abstract}

\section{Introduction}
\label{sec:intro}

Controlled text generation is the task of producing a sequence of coherent words based on given constraints. 
These constraints can range from simple attributes like tense, sentiment polarity and word-reordering \cite{hu2017toward,shen2017style,yang2018unsupervised} to more complex syntactic information. For example, given a sentence \textit{"The movie is awful!"} and a \textit{simple} constraint like flip sentiment to positive, a controlled text generator is expected to produce the sentence \textit{"The movie is fantastic!"}.  

These constraints are important in not only providing information about \textit{what to say} but also \textit{how to say it}. Without any constraint, the ubiquitous sequence-to-sequence neural models often tend to produce degenerate outputs and favour generic utterances \cite{vinyals2015neural,li2016diversity}.
While simple attributes are helpful in addressing \textit{what to say}, they provide very little information about \textit{how to say it}. Syntactic control over generation helps in filling this gap by providing that missing information.

\begin{table}[t]
    \scriptsize{\centering
        \begin{tabular}{m{4.4em}m{20.1em}}
            \toprule
            \textsc{Source} & -- \textit{how do i predict the stock market ?}\\
            \textsc{Exemplar} & -- \textit{can a brain transplant be done ?}\\
            \midrule
            \scpn & --  \textit{how can the stock and start ?}           \\
            \cgen & --  \textit{can the stock market actually happen ?}           \\
                \methodmain{} (Ours) & -- \textit{can i predict the stock market ?}\\
            \midrule
            \midrule
                \textsc{Source} & -- \textit{what are some of the mobile apps you ca n't live without and why ?}\\
                \textsc{Exemplar} & -- \textit{which is the best resume you have come across ?}\\
            \midrule
            \scpn & -- \textit{what are the best ways to lose weight ?}           \\
            \cgen & --  \textit{which is the best mobile app you ca n't ?}           \\
            \methodmain{} (Ours) & -- \textit{which is the best app you ca n't live without and why ?}\\
            \bottomrule
        \end{tabular}
        \caption{\label{tab:int} Sample syntactic paraphrases generated by \scpn{} \cite{iyyer2018adversarial}, \cgen{} \cite{chen2019controllable}, \methodmain{} (Ours). We observe that \methodmain{} is able to generate syntax conforming paraphrases without compromising much on relevance.}
}
\end{table}
\begin{figure*}[t]
    \centering
    \includegraphics[scale=0.44]{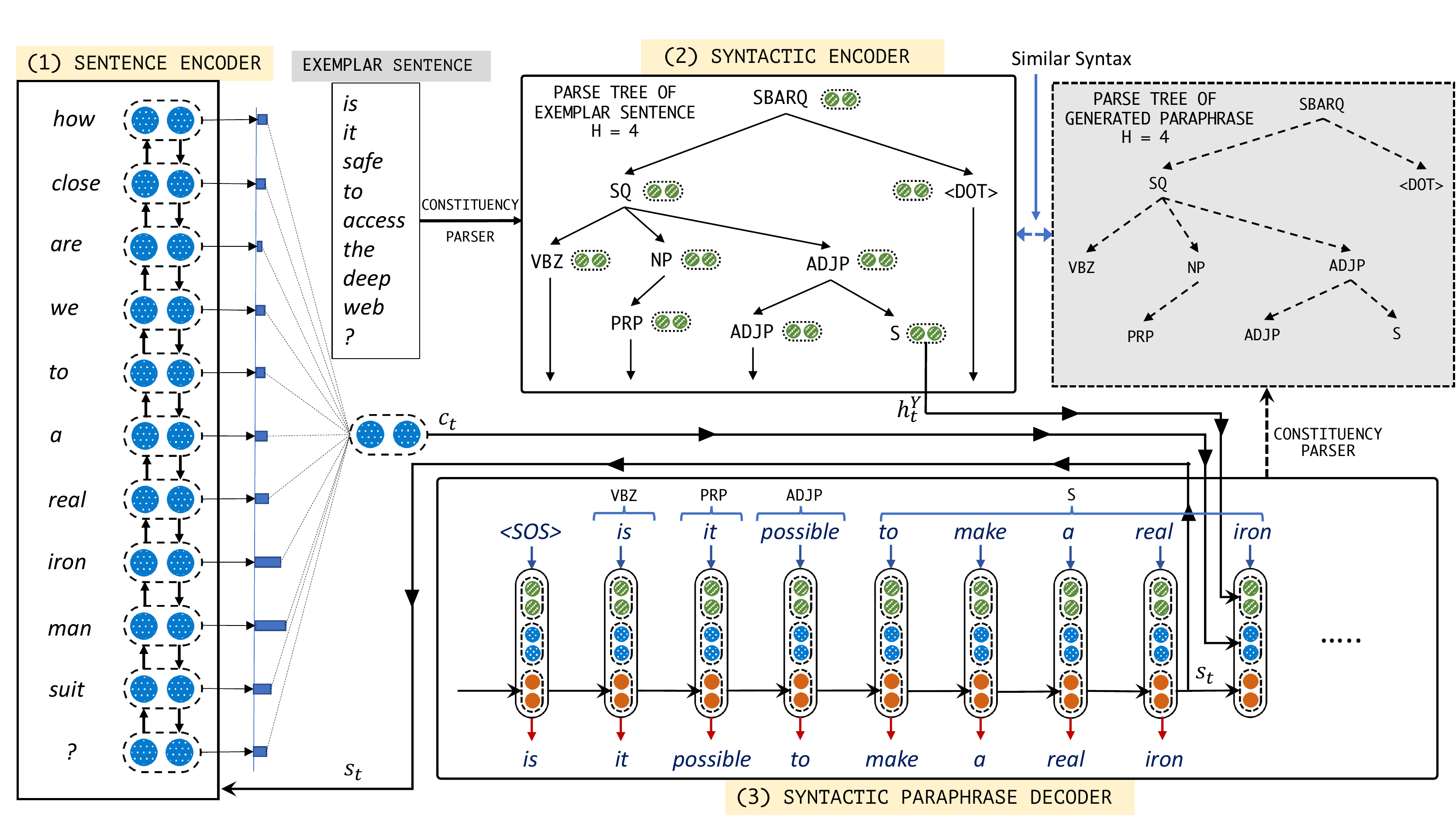}
    \caption{\label{fig:synpar} Architecture of \methodmain{} (proposed method). \methodmain{} aims to paraphrase an input sentence, while conforming to the syntax of an exemplar sentence (provided along with the input). 
    The input sentence is encoded using the Sentence Encoder (\refsec{subsec:sent-enc}) to obtain a semantic signal $c_t$. The Syntactic Encoder (\refsec{subsec:tree-enc}) takes a constituency parse tree (pruned at height $H$) of the exemplar sentence as an input, and produces representations for all the nodes in the pruned tree. 
    Once both of these are encoded, the Syntactic Paraphrase Decoder (\refsec{subsec:gendec}) uses pointer-generator network, and at each time step takes the semantic signal $c_t$, the decoder recurrent state $s_t$, embedding of the previous token and syntactic signal $h^Y_t$ 
    to generate a new token. Note that the syntactic signal remains the same for each token in a span (shown in figure above curly braces; please see \reffig{fig:constparse} for more details). The gray shaded region (not part of the model) illustrates a qualitative comparison of the exemplar syntax tree and the syntax tree obtained from the generated paraphrase. Please refer \refsec{sec:method} for details.}
\end{figure*}

Incorporating complex syntactic information has shown promising results in neural machine translation \cite{stahlberg2016syntactically,aharoni2017towards,yang2019latent}, data-to-text generation \cite{peng2019text}, abstractive text-summarization \cite{cao2018retrieve} and adversarial text generation \cite{iyyer2018adversarial}.
Additionally, recent work \cite{iyyer2018adversarial,kumar2019submodular} has shown that augmenting lexical and syntactical variations in the training set can help in building better performing and more robust models.

In this paper, we focus on the task of syntactically controlled paraphrase generation, i.e., given an input sentence and a syntactic exemplar, produce a sentence which conforms to the syntax of the exemplar while retaining the meaning of the original input sentence.
While syntactically controlled generation of paraphrases finds applications in multiple domains like data-augmentation and text passivization, we highlight its importance in the particular task of Text simplification. As pointed out in \citet{siddharthan2014survey}, depending on the literacy skill of an individual, certain syntactical forms of English sentences are easier to comprehend than others. As an example consider the following two sentences:
\begin{description}
    \setlength\itemsep{0.01em}
\item[S1] \textit{Because it is raining today, you should carry an umbrella.}
\item[S2] \textit{You should carry an umbrella today, because it is raining.}
\end{description}
Connectives that permit pre-posed adverbial clauses have been found to be difficult for third to fifth grade readers, even when the order of mention coincides with the causal (and temporal) order \cite{anderson1986conceptual, levy2003roots}. Hence, they prefer sentence \textbf{S2}. However, various other studies \cite{clark1968semantic, katz1968understanding, irwin1980effects} have suggested that for older school children, college students and adults, comprehension is better for the cause-effect presentation, hence sentence \textbf{S1}.
Thus, modifying a sentence, syntactically, would help in better comprehension based on literacy skills.

Prior work in syntactically controlled paraphrase generation addressed this task by conditioning the semantic input on either the features learnt from a linearized constituency-based parse tree \cite{iyyer2018adversarial}, or the \textit{latent} syntactic information \cite{chen2019controllable} learnt from exemplars through variational auto-encoders.
Linearizing parse trees, typically, result in loss of essential dependency information. On the other hand, as noted in \cite{shi2016string}, an auto-encoder based approach might not offer rich enough syntactic information as guaranteed by actual constituency parse trees.  
Moreover, as noted in \citet{chen2019controllable}, \scpn{} \cite{iyyer2018adversarial} and \cgen{} \cite{chen2019controllable} tend to generate sentences of the same length as the exemplar.  This is an undesirable characteristic because it often results in producing sentences that end abruptly, thereby compromising on grammaticality and semantics. Please see \reftbl{tab:int} for sample generations using each of the models.

To address these gaps, we propose \methodfull{} (\methodmain{}) which uses \emph{full} exemplar syntactic tree information. Additionally, our model provides an easy mechanism to incorporate different levels of syntactic control (granularity) based on the height of the tree being considered. The decoder in our framework is augmented with rich enough syntactical information to be able to produce syntax conforming sentences while not losing out on semantics and grammaticality.

The main contributions of this work are as follows:

\begin{enumerate}
    \setlength\itemsep{0.01em}
    \item We propose \methodfull{} (\methodmain{}), an end-to-end model to generate syntactically controlled paraphrases at different levels of granularity using a parsed exemplar.
    \item We provide a new decoding mechanism to incorporate syntactic information from the exemplar sentence's syntactic parse. 
    \item We provide a dataset formed from Quora Question Pairs \footnote{\href{https://www.kaggle.com/c/quora-question-pairs}{https://www.kaggle.com/c/quora-question-pairs}} for evaluating the models. We also perform extensive experiments to demonstrate the efficacy of our model using multiple automated metrics as well as human evaluations.
\end{enumerate}

\section{Related Work}
\label{sec:relatedwork}
\textbf{Controllable Text Generation} is an important problem in NLP 
which has received significant attention in recent times. Prior works include generating text using models conditioned on attributes like formality, sentiment or tense \cite{hu2017toward,shen2017style,yang2018unsupervised} as well as on syntactical templates \cite{iyyer2018adversarial,chen2019controllable}. These systems find applications in adversarial sample generation \cite{iyyer2018adversarial}, text summarization and table-to-text generation \cite{peng2019text}. While achieving state-of-the-art in their respective domains, these systems typically rely on a known finite set of attributes thereby making them quite restrictive in terms of the styles they can offer.

\noindent\textbf{Paraphrase generation}. While generation of paraphrases has been addressed in the past using traditional methods~\cite{mckeown1983paraphrasing,barzilay2003learning,quirk2004monolingual,hassan2007unt,zhao2008combining,madnani2010generating,wubben2010paraphrase}, they have recently been superseded by deep learning-based approaches \cite{prakash2016neural,gupta2018deep,li2019decomposable,li2018paraphrase,kumar2019submodular}. The primary task of all these methods \cite{prakash2016neural,gupta2018deep,li2018paraphrase} is to generate the most semantically similar sentence and they typically rely on beam search to obtain any kind of lexical diversity. \citet{kumar2019submodular} try to tackle the problem of achieving lexical, and limited syntactical diversity using submodular optimization but do not provide any syntactic control over the type of utterance that might be desired. These methods are therefore restrictive in terms of the syntactical diversity that they can offer.

\noindent \textbf{Controlled Paraphrase Generation}.  Our task is similar in spirit to \citet{iyyer2018adversarial,chen2019controllable}, which also deals with the task of syntactic paraphrase generation. However, the approach taken by them is different from ours in at least two aspects. Firstly, SCPN \cite{iyyer2018adversarial} uses attention \cite{bahdanau2014neural} based pointer-generator network \cite{see2017get} to encode input sentences and a linearised constituency tree to produce paraphrases. Due to the linearization of syntactic tree, a lot of dependency-based information is generally lost. Our model, instead, directly encodes the tree structure to produce a paraphrase. Secondly, the inference (or generation) process in SCPN is computationally very expensive, since it involves a two-stage generation process. In the first stage, they generate full parse trees from incomplete templates, and then from full parse trees to final generations. In contrast, the inference in our method involves a single-stage process, wherein our model takes as input a semantic source, a syntactic tree and the level of syntactic style that needs to be transferred, to obtain the generations.
Additionally, we also observed that the model does not perform well in low resource settings. This, again, can be attributed to the compounding implicit noise in the training due to linearised trees and generation of full linearised trees before obtaining the final paraphrases.

\citet{chen2019controllable} propose a syntactic exemplar-based method for controlled paraphrase generation using an approach based on latent variable probabilistic modeling, neural variational inference, and multi-task learning. This, in principle, is very similar to \citet{chen2019multi}. As opposed to our model which provides different levels of syntactic control of the exemplar-based generation, this approach is restrictive in terms of the flexibility it can offer.
Also, as noted in \citet{shi2016string}, an auto-encoder based approach might not offer rich enough syntactic information as offered by actual constituency parse trees. Additionally, VAEs \cite{kingma2014auto} are generally unstable and harder to train \cite{bowman2016generating,gupta2018deep} than seq2seq based approaches.

\section{\methodmain{}: Proposed Method}
\label{sec:method}
\begin{figure*}
\centering
\begin{subfigure}[t]{0.50\textwidth}
    \centering
    \includegraphics[scale=0.38]{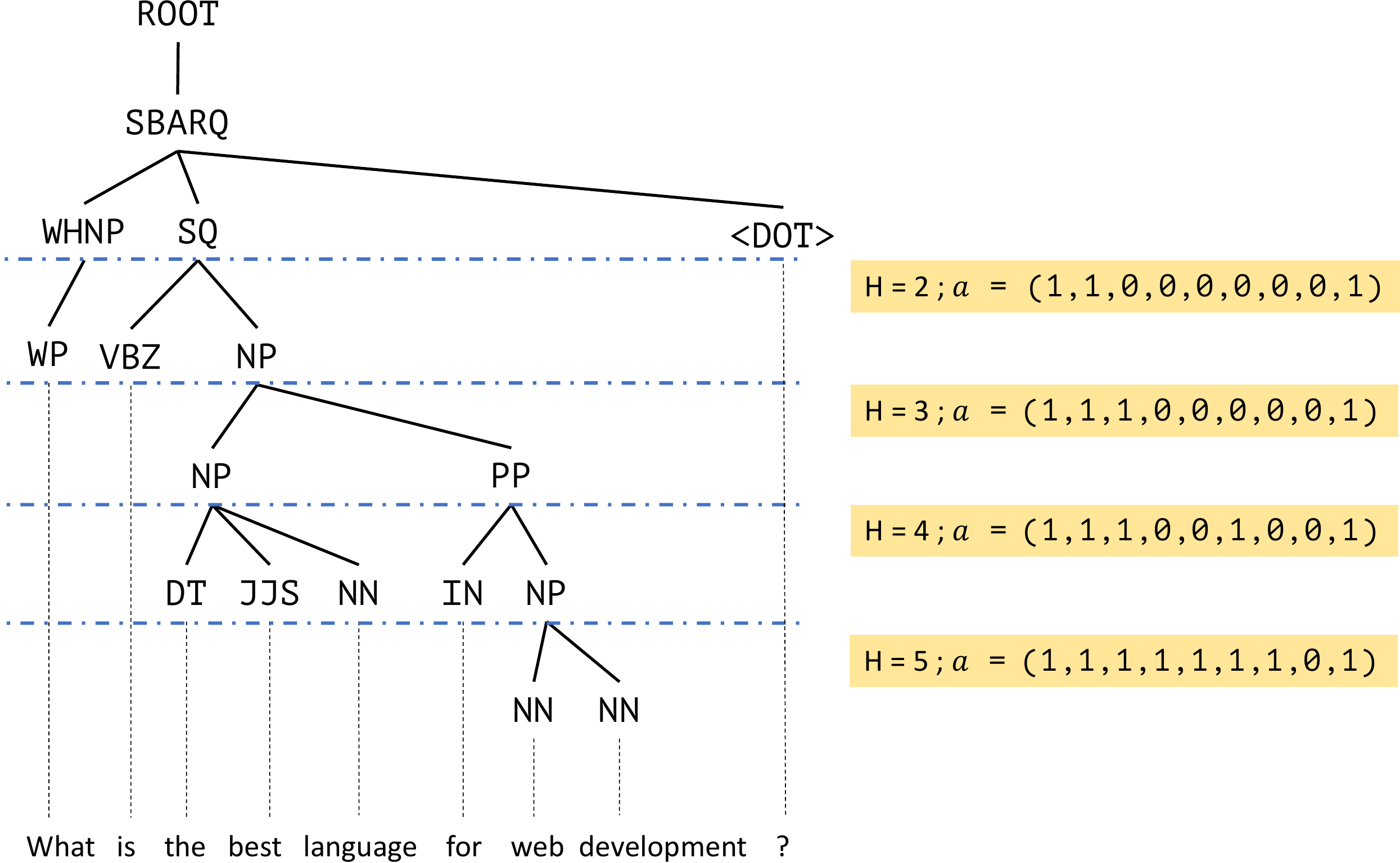}
    \caption{\label{fig:fullt}Full Constituency Parse Tree}
\end{subfigure}\hfill
\begin{subfigure}[t]{0.40\textwidth}
    \centering
    \includegraphics[scale=0.48]{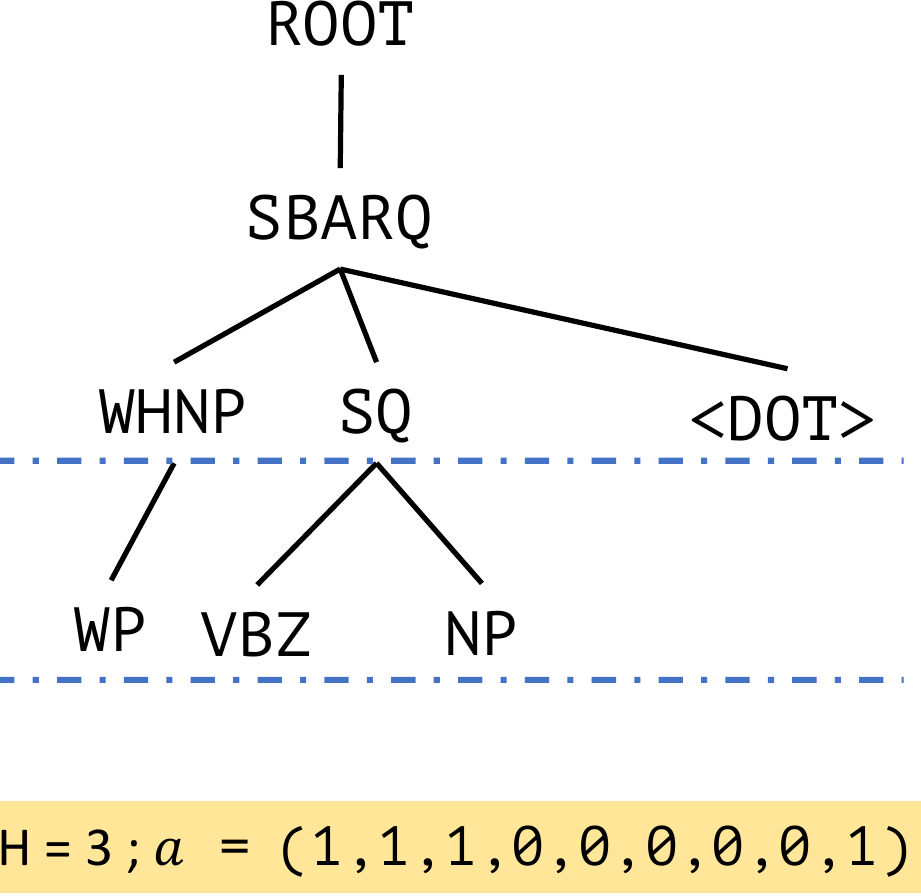}
    \caption{\label{fig:zoomt}Pruned tree at height H = 3}
\end{subfigure}
\caption{\label{fig:constparse} The constituency parse tree serves as an input to the syntactic encoder (\refsec{subsec:tree-enc}). The first step is to remove the leaf nodes which contain \textit{meaning representative tokens} (Here: What is the best language ...). $H$ denotes the height to which the tree can be pruned and is an input to the model. Figure (a) shows the full constituency parse tree annotated with vector $\boldsymbol{a}$ for different heights. Figure (b) shows the same tree pruned at height H = 3 with its corresponding $\boldsymbol{a}$ vector. The vector $\boldsymbol{a}$ serves as an \textit{signalling} vector (\refsec{subsubsec:syninfo}) which helps in deciding the syntactic signal to be passed on to the decoder. Please refer \refsec{sec:method} for details.}
\end{figure*}

In this section, we describe the inputs and various architectural components, essential for building \methodmain{}, an end-to-end trainable model. Our model, as shown in \reffig{fig:synpar}, comprises a sentence encoder (\ref{subsec:sent-enc}), syntactic tree encoder (\ref{subsec:tree-enc}), and a syntactic-paraphrase-decoder (\ref{subsec:gendec}).

\subsection{Inputs}
\label{subsec:inputs}
Given an input sentence $X$  and a syntactic exemplar $Y$, our goal is to generate a sentence $Z$ that conforms to the syntax of $Y$ while retaining the meaning of $X$.

While the semantic encoder (\refsec{subsec:sent-enc}) works on sequence of input tokens, the syntactic encoder (\refsec{subsec:tree-enc}) operates on constituency-based parse trees. We parse the syntactic exemplar $Y$\footnote{Obtained using the Stanford CoreNLP toolkit \cite{manning2014nlp}} to obtain its constituency-based parse tree.
The leaf nodes of the constituency-based parse tree consists of token for the sentence Y. These tokens, in some sense, carry the semantic information of sentence Y, which we do not need for generating paraphrases.
In order to prevent any \textit{meaning} propagation from exemplar sentence $Y$ into the generation, we remove these leaf/terminal nodes from its constituency parse. The tree thus obtained is denoted as $\mathcal{C}^Y$.

The syntactic encoder, additionally, takes as input $H$, which governs the level of syntactic control needed to be induced. The utility of H will be described in \refsec{subsec:tree-enc}.

\subsection{Semantic Encoder}
\label{subsec:sent-enc}

The semantic encoder, a multi-layered Gated Recurrent Unit (GRU), receives tokenized sentence $X = \{x_1, \ldots, x_{T_X}\}$ as input and computes the contextualized hidden state representation $h^{X}_t$ for each token using:
\begin{equation}
    h^{X}_t = \text{GRU}(h^{X}_{t-1}, e(x_t)),
    \label{eqn:hsent}
\end{equation}
where $e(x_t)$ represents the learnable embedding of the token $x_t$ and $t \in \{1, \ldots, T_X\}$ . Note that we use byte-pair encoding \cite{sennrich2016neural} for word/token segmentation.

\subsection{Syntactic Encoder}
\label{subsec:tree-enc}
This encoder provides the necessary syntactic guidance for the generation of paraphrases. Formally, let constituency tree $\mathcal{C}^Y = \{ \mathcal{V},\mathcal{E}, \mathcal{Y}\}$, where $\mathcal{V}$ is the set of nodes, $\mathcal{E}$ the set of edges and  $\mathcal{Y}$ the labels associated with each node.

We calculate the hidden-state representation $h^Y_v$ of each node $v \in \mathcal{V}$ using the hidden-state representation of its parent node $pa(v)$ and the embedding associated with its label $y_v$ as follows:
\begin{equation}
    h^Y_v = \text{GeLU}(W_{pa}h^Y_{pa(v)}+ W_ve(y_v) + b_v),
    \label{eqn:noderep}
\end{equation}
where $e(y_v)$ is the embedding of the node label $y_v$, and $W_{pa}, W_v, b_v$ are learnable parameters. This approach can be considered similar to TreeLSTM \cite{tai2015improved}. We use \texttt{GeLU} activation function \cite{hendrycks2016gaussian} rather than the standard \texttt{tanh} or \texttt{relu}, because of superior empirical performance.

As indicated in \refsec{subsec:inputs}, syntactic encoder takes as input the height $H$, which governs the level of syntactic control.
We randomly prune the tree $\mathcal{C}^Y$ to height $H \in$ $\{3, \ldots, H_{\texttt{max}}\} $, where $H_{\texttt{max}}$ is the height of the full constituency tree $\mathcal{C}^Y$. As an example, in \reffig{fig:zoomt}, we prune the constituency-based parse tree of the exemplar sentence, to height $H = 3$. The leaf nodes for this tree have the labels \texttt{WP, VBZ, NP} and \texttt{<DOT>}. While we calculate the hidden-state representation of all the nodes, only the terminal nodes are responsible for providing the syntactic signal to the decoder (\refsec{subsec:gendec}).

We maintain a queue $\mathbb{L}_H^Y$ of such terminal node representations where elements are inserted from left to right for a given H.
Specifically, for the particular example given in \reffig{fig:zoomt},
$$\mathbb{L}_H^Y = [h^Y_{\texttt{WP}},h^Y_{\texttt{VBZ}},h^Y_{\texttt{NP}},h^Y_{\texttt{<DOT>}}]$$

\noindent We emphasize the fact that the length of the queue $|\mathbb{L}_H^Y|$ is a function of height $H$.

\subsection{Syntactic Paraphrase Decoder}
\label{subsec:gendec}
Having obtained the semantic and syntactic representations, the decoder is tasked with the generation of syntactic paraphrases.
This can be modeled as finding the best $Z = Z^*$ that maximizes the probability $\mathbb{P}(Z|X, Y)$, which can further be factorized as:
\begin{equation}
    Z^* = \argmax_z \prod^{T_Z}_{t=1}(z_t | z_1, \ldots, z_{t-1}, X, Y),
\end{equation}
where $T_Z$ is the maximum length up to which decoding is required.

In the subsequent sections, we use $t$ to denote the decoder time step.

\subsubsection{Using Semantic Information}
\label{subsubsec:seminfo}
At each decoder time step $t$, the attention distribution $\alpha^t$ is calculated over the encoder hidden states $h^X_i$, obtained using \refeqn{eqn:hsent}, as:
\begin{equation}
    \begin{split}
        e^t_i = v^\intercal\text{tanh}(W_hh^X_i + W_ss_t + b_{\text{attn}})\\
        \alpha^t = \text{softmax}(e^t),
    \end{split}
    \label{eqn:attndist}
\end{equation}
where $s_t$ is the decoder cell-state and $v, W_h, W_s, b_{\text{attn}}$ are learnable parameters.

The attention distribution provides a way to jointly-align and train sequence to sequence models by producing a weighted sum of the semantic encoder hidden states, known as context-vector $c_t$ given by:
\begin{equation}
    c_t = \sum_i \alpha^t_ih^X_i
\end{equation}

\noindent $c_t$ serves as the semantic signal which is essential for generating meaning preserving sentences.

\subsubsection{Using Syntactic Information}
\label{subsubsec:syninfo}
During \textit{training}, each terminal node in the tree $\mathcal{C}^Y$, pruned at H, is equipped with information about the span of words it needs to generate.
At each time step $t$, only \textit{one} terminal node representation $h^Y_v \in \mathbb{L}^Y_H$ is responsible for providing the syntactic signal which we call $h^Y_t$. This hidden-state representation to be used is governed through an \textit{signalling} vector $\boldsymbol{a} = (a_1, \ldots, a_{T_z})$, where each $a_i \in \{0, 1\}$. 0 indicates that the decoder should keep on using the same hidden-representation $h^Y_v \in \mathbb{L}^Y_H$ that is currently being used, and 1 indicates that the next element (hidden-representation) in the queue $\mathbb{L}^Y_H$ should be used for decoding.

The utility of $\boldsymbol{a}$ can be best understood through \reffig{fig:zoomt}. Consider the syntactic tree pruned at height $H=3$. For this example, $$\mathbb{L}_H^Y = [h^Y_{\texttt{WP}}, h^Y_{\texttt{VBZ}}, h^Y_{\texttt{NP}}, h^Y_{\texttt{<DOT>}}]$$ and $$\boldsymbol{a} = (1, 1, 1, 0, 0, 0, 0, 0, 1)$$

$a_i = 1$ provides a signal to pop an element from the queue $\mathbb{L}^Y_H$ while $a_i = 0$ provides a signal to keep on using the last popped element. This element is then used to guide the decoder \textit{syntactically} by providing a signal in the form of hidden-state representation (\refeqn{eqn:p_voc}).

Specifically, in this example, the $a_1 = 1$ signals $\mathbb{L}^Y_H$ to pop $h^Y_{\texttt{WP}}$ to provide syntactic guidance to the decoder for generating the first token. $a_2 = 1$ signals $\mathbb{L}^Y_H$ to pop $h^Y_{\texttt{VBZ}}$ to provide syntactic guidance to the decoder for generating the second token. $a_3 = 1$ helps in obtaining $h^Y_{\texttt{NP}}$ from $\mathbb{L}^Y_H$ to provide guidance to generate the third token. As described earlier, $a_4, \ldots, a_8 = 0$ indicate that the same representation $h^Y_{\texttt{NP}}$ should be used for syntactically guiding tokens $z_4, \ldots, z_8$. Finally $a_9 = 1$ helps in retrieving $h^Y_{\texttt{<DOT>}}$ for guiding decoder to generate token $z_9$.
Note that $|\mathbb{L}_H^Y| = \sum_{i=1}^{T_z} a_i$

While $\boldsymbol{a}$ is provided to the model during training, this information might not be available during inference. Providing $\boldsymbol{a}$ during generation makes the model restrictive and might result in producing ungrammatical sentences. \methodmain{} is tasked to learn a proxy for the \textit{signalling} vector $\boldsymbol{a}$, using \textit{transition probability vector} $\boldsymbol{p}$.

At each time step $t$, we calculate $p_t \in (0, 1)$ which determines the probability of changing the syntactic signal using:
\begin{equation}
    \label{eqn:p_bop}
    p_t = \sigma (W_{\text{bop}}([c_t; h^Y_t; s_t; e(z^\prime_t)]) + b_{\text{bop}}),
\end{equation}

\begin{equation}
    h^Y_{t+1} =
    \begin{cases}
      h^Y_t & p_t < 0.5 \\
      \texttt{pop}(\mathbb{L}_H^Y)& \text{otherwise}
   \end{cases}
   \label{eqn:bop}
\end{equation}
where \texttt{pop} removes and returns the next element in the queue, $s_t$ is the decoder state, and $e(z^\prime_t)$ is the embedding of the input token at time $t$ during decoding.

\subsubsection{Overall}
The semantic signal $c_t$, together with decoder state $s_t$, embedding of the input token $e(z^\prime_{t})$ and the syntactic signal $h^Y_t$ is fed through a GRU followed by softmax of the output to produce a vocabulary distribution as:
\begin{equation}
    \label{eqn:p_voc}
    \mathbb{P}_{\text{vocab}} = \text{softmax}(W([c_t;h^Y_t;s_t; e(z^\prime_{t})]) + b),
\end{equation}
where $[;]$ represents concatenation of constituent elements, and $W, b$ are trainable parameters.

We augment this with the copying mechanism \cite{vinyals2015pointer} as in the pointer-generator network \cite{see2017get}. Usage of such a mechanism offers a probability distribution over the extended vocabulary (the union of vocabulary words and words present in the source sentence) as follows:
\begin{equation}
    \label{eqn:p_voc_n_copy}
    \begin{split}
        \mathbb{P}(z) = p_{\text{gen}}\mathbb{P}_{\text{vocab}}(z) + (1-p_{\text{gen}})\sum_{i:z_i = z}\alpha^t_i \\
        p_{\text{gen}} = \sigma(w_{c}^\intercal c_t + w_s^\intercal s_t + w_x^\intercal e(z^\prime_t) + b_{gen})
    \end{split}
\end{equation}
where $w_{c}, w_s, w_x\, \text{and}\,  b_{gen}$ are learnable parameters, $e(z^\prime_t)$ is the input token embedding to the decoder at time step $t$ and $\alpha_i^t$ is the element corresponding to the $i^{th}$ co-ordinate in the attention distribution as defined in \refeqn{eqn:attndist}

The overall objective can be obtained by taking negative log-likelihood of the distributions obtained in \refeqn{eqn:p_bop} and \refeqn{eqn:p_voc_n_copy}.
\begin{equation}
    \begin{split}
        \mathcal{L} =&- \frac{1}{T}\sum_{t=0}^T [\log \mathbb{P}(z^*_t)  \\
                    &+ a_t\log(p_t) \\
                    &+ (1-a_t) \log (1-p_t)]
    \end{split}
\end{equation}
where $a_t$ is the $t^{th}$ element of the vector $\boldsymbol{a}$.

\section{Experiments}
\label{sec:experiments}

Our experiments are geared towards answering the following questions:
\begin{description}
    \setlength\itemsep{0.01em}
    \item[Q1.] Is \methodmain{} able to generate syntax conforming sentences without losing out on meaning? (\refsec{subsec:sempres}, \ref{subsec:qual_analysis})
    \item[Q2.] What level of syntactic control does \methodmain{} offer? (\refsec{subsec:syngran}, \ref{subsec:posbleu}, \ref{subsec:syngran})
    \item[Q3.] How does \methodmain{} compare against prior models, qualitatively? (\refsec{subsec:qual_analysis})
    \item[Q4.] Are the improvements achieved by \methodmain{} statistically significant? (\refsec{subsec:sempres})
\end{description}
Based on these questions, we outline the methods compared (\refsec{subsec:baseline}), along with the datasets (\refsec{subsec:dataset}) used, evaluation criteria (\refsec{subsec:evaluation}) and the experimental setup (\refsec{subsec:exptsetup}).

\subsection{Methods Compared}
\label{subsec:baseline}
As in \citet{chen2019controllable}, we first highlight the results of the two direct return-input baselines.
\begin{enumerate}
    \setlength\itemsep{0.01em}
    \item \textbf{Source-as-Output}: Baseline where the output is the semantic input.
    \item \textbf{Exemplar-as-Output}: Baseline where the output is the syntactic exemplar.
\end{enumerate}
We compare the following competitive methods:
\begin{enumerate}
    \setcounter{enumi}{2}
    \setlength\itemsep{0.01em}
\item \textbf{\scpn{}} \cite{iyyer2018adversarial} is a sequence-to-sequence based model comprising two encoders built with LSTM \cite{hochreiter1997long} to encode semantics and syntax respectively. Once the encoding is obtained, it serves as an input to the LSTM based decoder which is augmented with soft-attention \cite{bahdanau2014neural} over encoded states as well as a copying mechanism \cite{see2017get} to deal with out-of-vocabulary tokens.  \footnote{Note that the results for \scpn{} differ from the ones shown in \cite{iyyer2018adversarial}. This is because the dataset used in \cite{iyyer2018adversarial} is atleast 50 times larger than the largest dataset (ParaNMT-small) in this work.}
\item \textbf{\cgen{}} \cite{chen2019controllable} is a VAE \cite{kingma2014auto} model with two encoders to project semantic input and syntactic input to a latent space. They obtain a syntactic embedding from one encoder, using a standard Gaussian prior. To obtain the semantic representation, they use von Mises-Fisher prior, which can be thought of as a Gaussian distribution on a hypersphere. They train the model using a multi-task paradigm, incorporating paraphrase generation loss and word position loss. We considered their best model, VGVAE + LC + WN + WPL, which incorporates the above objectives.
    \item \textbf{\methodmain{} (\refsec{sec:method}) } is a sequence-and-tree-to-sequence based model which encodes semantics and tree-level syntax to produce paraphrases.
    It uses a GRU \cite{chung2014empirical} based decoder with soft-attention on semantic encodings and a \textit{begin of phrase} (bop) gate to select a leaf node in the exemplar syntax tree.
    We compare the following two variants of \methodmain{}:\newline
    \textbf{(a) \methodmain{}-F : } Uses full constituency parse tree information of the exemplar for generating paraphrases.\newline
    \textbf{(a) \methodmain{}-R : } \methodmain{} can produce multiple paraphrases by pruning the exemplar tree at various heights. This variant first generates 5 candidate generations, corresponding to 5 different heights of the exemplar tree namely $\{H_\texttt{max}, H_\texttt{max}-1, H_\texttt{max}-2, H_\texttt{max}-3, H_\texttt{max}-4\}$, for each (source, exemplar) pair. From these candidates, the one the highest ROUGE-1 score with the source sentence, is selected as the final generation.
\end{enumerate}

Note that, except for the return-input baselines, all methods use beam search during inference.

\subsection{Datasets}
\label{subsec:dataset}
We train the models and evaluate them on the following datasets:

\noindent \textbf{(1) ParaNMT-small} \cite{chen2019controllable} contains 500K sentence-paraphrase pairs for training, and 1300 manually labeled sentence-exemplar-reference which is further split into 800 test data points and 500 dev. data points respectively.

\noindent As in \citet{chen2019controllable}, our model uses only (sentence, paraphrase) during training. The paraphrase itself serves as the exemplar input during training.

        This dataset is a subset of the original ParaNMT-50M dataset \cite{wieting2018paranmt}. ParaNMT-50M is a data set generated automatically through backtranslation of original English sentences. It is inherently noisy due to imperfect neural machine translation quality with many sentences being non-grammatical and some even being non-English sentences. Because of such noisy data points, it is optimistic to assume that the corresponding constituency parse tree would be well aligned. To that end, we propose to use the following additional dataset which is more well-formed and has more human intervention than the ParaNMT-50M dataset.\newline

\noindent \textbf{(2) QQP-Pos:} The original Quora Question Pairs (QQP) dataset contains about 400K sentence pairs labeled positive if they are duplicates of each other and negative otherwise.
    The dataset is composed of about 150K positive and 250K negative pairs. We select those positive pairs which contain both sentences with a maximum token length of 30, leaving us with \textasciitilde146K pairs. We call this dataset as QQP-Pos.

    Similar to ParaNMT-small, we use only the sentence-paraphrase pairs as training set and sentence-exemplar-reference triples for testing and validation.
    We randomly choose 140K sentence-paraphrase pairs as the training set $\mathbb{T}_{train}$, and the remaining 6K pairs $\mathbb{T}_{eval}$ are used to form the evaluation set $\mathbb{E}$. Additionally, let $\mathbb{T}_{eset} = \bigcup\{\{X,Z\}: (X,Z) \in \mathbb{T}_{eval}\}$. Note that $\mathbb{T}_{eset}$ is a set of sentences while $\mathbb{T}_{eval}$ is a set of sentence-paraphrase pairs.

    \noindent Let $\mathbb{E} = \phi$ be the initial evaluation set.
    For selecting exemplar for each \textit{each sentence-paraphrase pair} $(X, Z) \in \mathbb{T}_{eval}$, we adopt the following procedure:
    \begin{description}
        \setlength\itemsep{0.01em}
    \item[Step 1:] For a given $(X, Z) \in \mathbb{T}_{eval}$, construct an exemplar candidate set $\mathbb{C} = \mathbb{T}_{eset} - \{X, Z\}$. $|\mathbb{C}| \approx 12,000$. 
    \item[Step 2:] Retain only those sentences $C \in \mathbb{C}$ whose sentence length (= number of tokens) differ by at most 2 when compared to the paraphrase $Z$. This is done since sentences with similar constituency-based parse tree structures tend to have similar token lengths.
    \item[Step 3:] Remove those candidates $C \in \mathbb{C}$, which are very similar to the source sentence $X$, i.e. $\texttt{BLEU}(X, C) > 0.6$.
    \item[Step 4:] From the remaining instances in $\mathbb{C}$, choose that sentence $C$ as the exemplar $Y$ which has the least Tree-Edit distance with the paraphrase $Z$ of the selected pair i.e. $Y = \underset{C \in \mathbb{C}}{\mathrm{argmin}}\; \texttt{TED}(Z, C)$. This ensures that the constituency-based parse tree of the exemplar $Y$ is quite similar to that of $Z$, in terms of Tree-Edit distance.
    \item[Step 5:] $\mathbb{E} := \mathbb{E} \cup	(X, Y, Z)$
    \item[Step 6:] Repeat procedure for all other pairs in $\mathbb{T}_{eval}$.
\end{description}

    From the obtained evaluation set $\mathbb{E}$, we randomly choose 3K triplets for the test set $\mathbb{T}_{test}$, and remaining 3K for the validation set $\mathbb{V}$.

\begin{table*}[th]
    \scriptsize{\centering
        \begin{tabular}{m{10.5em}P{3em}P{4.0em}P{5.1em}P{5.1em}P{5.2em}P{3.8em}P{3.8em}P{3em}}
            \toprule
            \multicolumn{9}{c}{\textbf{QQP-Pos}}\\
            \toprule
            \textbf{Model} & \textbf{BLEU}$\uparrow$ & \textbf{METEOR}$\uparrow$ & \textbf{ROUGE-1}$\uparrow$ & \textbf{ROUGE-2}$\uparrow$ & \textbf{ROUGE-L}$\uparrow$ & \textbf{TED-R}$\downarrow$ & \textbf{TED-E}$\downarrow$ & \textbf{PDS}$\uparrow$ \\ 
            \midrule
            Source-as-Output & 17.2 & 31.1  & 51.9 & 26.2 & 52.9 & 16.2& 16.6 & 99.8 \\
            Exemplar-as-Output& 16.8 & 17.6 & 38.2 & 20.5 & 43.2 & 4.8& 0.0 & 10.7 \\
            \midrule
            \scpn{} \cite{iyyer2018adversarial} & 15.6 & 19.6 & 40.6 & 20.5 & 44.6 & 9.1 &8.0& 27.0 \\
            \cgen{} \cite{chen2019controllable} & 34.9 & 37.4 & 62.6 & 42.7 & 65.4  & 6.7& 6.0 & 65.4\\
            \midrule
                \methodmain{}-F & \textbf{36.7} &  \textbf{39.8} & \textbf{66.9} & \textbf{45.0} & \textbf{69.6} & \textbf{4.8} &\textbf{1.8}& \textbf{75.0}\\
               \methodmain{}-R& \fbox{38.0} & \fbox{41.3} & \fbox{68.1} & \fbox{45.7} & \fbox{70.2} & 6.8 & 5.9 & \fbox{87.7}\\
            \toprule
            \toprule
            \multicolumn{9}{c}{\textbf{ParaNMT-small}}\\
            \midrule
                Source-as-Output & 18.5 & 28.8 & 50.6 & 23.1 & 47.7 & 12.0 & 13.0& 99.0 \\
                Exemplar-as-Output & 3.3 & 12.1 & 24.4 & 7.5 & 29.1  & 5.9 & 0.0 & 14.0\\
            \midrule
                \scpn{} \cite{iyyer2018adversarial}  & 6.4 & 14.6 & 30.3 & 11.2 & 34.6 & 6.2 & \textbf{1.4} & 15.4 \\
                \cgen{} \cite{chen2019controllable} & 13.6 & 24.8 & 44.8 & 21.0 & 48.3  & 6.7 & 3.3 &70.2\\
            \midrule
                \methodmain{}-F & \textbf{15.3} & \textbf{25.9} & \textbf{46.6} & \underline{21.8} & \textbf{49.7} & \textbf{6.1} &\textbf{1.4} & \textbf{76.6}\\ 
                \methodmain{}-R & \fbox{16.4} & \fbox{27.2} & \fbox{49.6} & \fbox{22.9} & \fbox{50.5} & 8.7& 7.0 & \fbox{83.5}\\
            \bottomrule
        \end{tabular}
        \caption{\label{tab:synpar_res}Results on QQP and ParaNMT-small dataset.
        Higher$\uparrow$ BLEU, METEOR, ROUGE and PDS is better whereas lower$\downarrow$ TED score is better. \methodmain{}-R selects the best candidate out of many, resulting in performance boost for semantic preservation (shown in box). We bold the statistically significant results of \methodmain{}-F, only, for a fair comparison with the baselines. Note that Source-as-Output, and Exemplar-as-Output are only dataset quality indicators and not the competitive baselines. Please see \refsec{sec:results} for details.}
    }
\end{table*}

\subsection{Evaluation}
\label{subsec:evaluation}
It should be noted that there is no single fully-reliable metric for evaluating syntactic paraphrase generation. Therefore, we evaluate on the following metrics to showcase the efficacy of syntactic paraphrasing models.
\begin{enumerate}
    \setlength\itemsep{0.01em}
    \item \textbf{Automated Evaluation}. \\
    \textbf{(i) Alignment based metrics:} We compute BLEU \cite{papineni2002bleu}, METEOR \cite{banerjee2005meteor}, ROUGE-1, ROUGE-2, ROUGE-L \cite{lin2004rouge} scores between the generated and the reference paraphrases in the test set.

    \textbf{(ii) Syntactic Transfer}: We evaluate the syntactic transfer using Tree-edit distance \cite{zhang1989simple} between the parse trees of:
        \begin{enumerate}
            \setlength\itemsep{0.01em}
            \item the generated and the syntactic exemplar in the test set - \textbf{TED-E}
        \item the generated and the reference paraphrase in the test set - \textbf{TED-R}
        \end{enumerate}
    \textbf{(iii) Model-based evaluation:} Since our goal is to generate paraphrases of the input sentences, we need some measure to determine if the generations indeed convey the same meaning as the original text.
    To achieve this, we adopt a model-based evaluation metric as used by \citet{shen2017style} for Text Style Transfer and \citet{isola2017image} for Image Transfer. Specifically, classifiers are trained on the task of Paraphrase Detection and then used as Oracles to evaluate the generations of our model and the baselines.
    We fine-tune two RoBERTa \cite{liu2019roberta} based sentence pair classifiers, one on Quora Question Pairs (\textit{Classifier-1}) and other on ParaNMT + PAWS\footnote{Since the ParaNMT dataset only contains paraphrase pairs, we augment it with PAWS \cite{zhang2019paws} dataset to acquire negative samples.} datasets (\textit{Classifier-2}) which achieve accuracies of 90.2\% and 94.0\% on their respective test sets\footnote{Since the test set of QQP is not public, the 90.2\% number was computed on the available dev set (not used for model selection)}.

    Once trained, we use \textit{Classifier-1} to evaluate generations on QQP-Pos and \textit{Classifier-2} on ParaNMT-small.

    We first generate syntactic paraphrases using all the models (\refsec{subsec:baseline}) on the test splits of QQP-Pos and ParaNMT-small datasets. We then pair the source sentence with their corresponding generated paraphrases and send them as input to the classifiers.
    The Paraphrase Detection score, denoted as \textit{PDS} in \reftbl{tab:synpar_res}, is defined as,
    the ratio of the number of generations predicted as paraphrases of their corresponding source sentences by the classifier to the total number of generations.

\item \textbf{Human Evaluation.}\\ While \texttt{TED} is sufficient to highlight syntactic transfer, there has been some scepticism regarding automated metrics for paraphrase quality \cite{reiter2018bleu}. To address this issue, we perform human evaluation on 100 randomly selected data points from the test set. In the evaluation, 3 judges (non-researchers proficient in the English language) were asked to assign scores to generated sentences based on the semantic similarity with the given source sentence. The annotators were shown a source sentence and the corresponding outputs of the systems in random order. The scores ranged from 1 (doesn't capture meaning at all) to 4 (perfectly captures the meaning of the source sentence).

\end{enumerate}

\subsection{Setup}
\label{subsec:exptsetup}

\noindent \textbf{(a) Pre-processing}. Since our model needs access to constituency parse trees, we tokenize and parse all our data points using the fully parallelizable Stanford CoreNLP Parser \cite{manning2014nlp} to obtain their respective parse trees. This is done prior to training in order to prevent any additional computational costs that might be incurred because of repeated parsing of the same data points during different epochs.
\begin{table*}[th]
    \scriptsize{\centering
        \begin{tabular}{m{13em}m{40em}}
            \toprule
            Source  & \textit{what should be done to get rid of laziness ?}           \\
            Template Exemplar & \textit{how can i manage my anger ?}           \\
            \midrule
            \scpn{} \cite{iyyer2018adversarial}  & \textit{how can i get rid ?}           \\
            \cgen{} \cite{chen2019controllable} & \textit{how can i get rid of ?}           \\
                \methodmain{}-F (Ours) & \textit{how can i stop my laziness ?}\\
            \methodmain{}-R (Ours) & \textit{how do i get rid of laziness ?}           \\
            \midrule
            \midrule
            Source  & \textit{what books should entrepreneurs read on entrepreneurship ?}\\ 
                Template Exemplar  & \textit{what is the best programming language for beginners to learn ?}\\ 
            \midrule
            \scpn{} \cite{iyyer2018adversarial}  & \textit{what are the best books books to read to read ?}\\
            \cgen{} \cite{chen2019controllable}  & \textit{what 's the best book for entrepreneurs read to entrepreneurs ?}\\
            \methodmain{}-F (Ours) & \textit{what is a best book idea that entrepreneurs to read ?}\\
            \methodmain{}-R (Ours) & \textit{what is a good book that entrepreneurs should read ?}           \\
            \midrule
            \midrule
            Source  & \textit{how do i get on the board of directors of a non profit or a for profit organisation ?}\\
            Template Exemplar & \textit{what is the best way to travel around the world for free ?}           \\
            \midrule
            \scpn{} \cite{iyyer2018adversarial} & \textit{what is the best way to prepare for a girl of a ?}           \\
            \cgen{} \cite{chen2019controllable}  & \textit{what is the best way to get a non profit on directors ?}           \\
            \methodmain{}-F (Ours) & \textit{what is the best way to get on the board of directors ?}\\
            \methodmain{}-R (Ours) & \textit{what is the best way to get on the board of directors of a non profit or a for profit organisation ?}           \\
            \bottomrule
        \end{tabular}
        \caption{\label{tab:generations}Sample generations of the competitive models. Please refer to \refsec{subsec:discussion} for details}
    }
\end{table*}

\noindent \textbf{(b) Implementation details}. We train both our models using the Adam Optimizer \cite{kingma2014adam} with an initial learning rate of 7e-5. We use a bidirectional 3-layered GRU for encoding the tokenized semantic input and a standard pointer-generator network with GRU for decoding. The token embedding is learnable with dimension 300. To reduce the training complexity of the model, the maximum sequence length is kept at 60.  The vocabulary size is kept at 24K for QQP and 50K for ParaNMT-small.

\methodmain{} needs access to the level of syntactic granularity for decoding, depicted as $H$ in \reffig{fig:constparse}. During \textit{training}, we keep on varying it randomly from 3 to $H_{\texttt{max}}$, changing it with each training epoch. This ensures that our model is able to generalize because of an implicit regularization attained using this procedure. At each time-step of the decoding process, we keep a teacher forcing ratio of 0.9.

\section{Results}
\label{sec:results}

\subsection{Semantic Preservation and Syntactic transfer}
\label{subsec:sempres}

\textbf{1. Automated Metrics: } As can be observed in \reftbl{tab:synpar_res}, our method(s) (\methodmain{}-F/R (\refsec{subsec:baseline})) are able to outperform the existing baselines on both the datasets. Source-as-Output is independent of the exemplar sentence being used and since a sentence is a paraphrase of itself, the \emph{paraphrastic scores} are generally high while the \textit{syntactic scores} are below par. An opposite is true for Exemplar-as-Output. These baselines also serve as \emph{dataset quality} indicators. It can be seen that source is semantically similar while being syntactically different from target sentence whereas the opposite is true when exemplar is compared to target sentences. Additionally, source sentences are syntactically and semantically different from exemplar sentences as can be observed from TED-E and PDS scores. This helps in showing that the dataset has rich enough syntactic diversity to learn from.

Through TED-E scores it can be seen that \methodmain{}-F is able to adhere to the syntax of the exemplar template to a much larger degree than the baseline models.
This verifies that our model is able to generate meaning preserving sentences while conforming to the syntax of the exemplars when measured using standard metrics.

It can also be seen that \methodmain{}-R tends to perform better than \methodmain{}-F in terms of \emph{paraphrastic scores} while taking a hit on the \emph{syntactic scores}. This makes sense, intuitively, because in some cases \methodmain{}-R tends to select lower $H$ values for syntactic granularity. This can also be observed from the example given in \reftbl{tab:heightgen} where $H=6$ is more favourable than $H=7$, because of better meaning retention.

  Although \cgen{} performs close to our model in terms of BLEU, ROUGE and METEOR scores on ParaNMT-small dataset, its PDS is still much lower than that of our model, suggesting that our model is better at capturing the original meaning of the source sentence.
In order to show that the results are not coincidental, we test the statistical significance of our model. We follow the non-parametric Pitman's permutation test \cite{dror2018hitchhikers} and observe that our model is statistically significant when the significance level ($\alpha$) is taken to be 0.05. Note that this holds true for all metric on both the datasets except ROUGE-2 on ParaNMT-small.\newline

\begin{table}[h!]
    \scriptsize{\centering
        \begin{tabular}{m{7em}m{3em}m{3em}m{3.5em}m{3.5em}}
            \toprule
            & \scpn{} & \cgen{} & \methodmain{}-F & \methodmain{}-R\\
            \midrule
            \textbf{QQP-Pos} & 1.63 & 2.47 & \textbf{2.70} & \fbox{2.99} \\
            \midrule
            \textbf{ParaNMT-small} & 1.24 & 1.89 &  \textbf{2.07} & \fbox{2.26} \\
            \bottomrule
        \end{tabular}
        \caption{\label{tab:human_eval}A comparison of human evaluation scores for comparing quality of paraphrases generated using all models. Higher score is better. Please refer to \refsec{subsec:sempres} for details. }
    }
\end{table}
\begin{table*}[th]
    \scriptsize{\centering
        \begin{tabular}{m{25em}m{29em}}
            \toprule
            \multicolumn{2}{c}{\textbf{SOURCE : }\textit{how do i develop my career in software ?}} \\
            \midrule
            \textbf{SYNTACTIC EXEMPLAR} & \textbf{\methodmain{} GENERATIONS}\\
            \midrule
            \textit{how can i get a domain for free ?} & \textit{how can i develop a career in software ?}\\
            \midrule
            \textit{what is the best way to register a company ?} & \textit{what is the best way to develop career in software ?}\\ 
            \midrule
            \textit{what are good places to visit in new york ?} & \textit{what are good ways to develop my career in software ?}\\
            \midrule
            \textit{can i make 800,000 a month betting on horses ?} & \textit{can i develop my career in software ?}\\
            \midrule
            \textit{what is chromosomal mutation ? what are some examples ?} & \textit{what is good career ? what are some of the ways to develop my career in software ?}\\
            \midrule
            \textit{is delivery free on quikr ?} & \textit{is career useful in software ?}\\
            \midrule
            \textit{is it possible to mute a question on quora ?} & \textit{is it possible to develop my career in software ?}\\
            \bottomrule
        \end{tabular}
        \caption{\label{tab:tempgen} Sample \methodmain{}-R generations with a single source sentence and multiple syntactic exemplars. Please refer to \refsec{subsec:qual_analysis} for details.}
    }
\end{table*}

\noindent  \textbf{2. Human Evaluation: } \reftbl{tab:human_eval} shows the results of human assessment. It can be seen that annotators, generally tend to rate \methodmain{}-F and \methodmain{}-R (\refsec{subsec:baseline}) higher than the baseline models, thereby highlighting the efficacy of our models.  This evaluation additionally shows that automated metrics are somewhat consistent with the human evaluation scores.

\subsection{Syntactic Control}
\label{subsec:syngran}

\textbf{1. Syntactical Granularity :}
Our model can work with different levels of granularity for the exemplar syntax, i.e.,  different tree heights of the exemplar tree can be used for decoding the output.

 \begin{table}[th]
    \scriptsize{\centering
        \begin{tabular}{m{2.2em}m{22.1em}}
            \toprule
            S & \textit{what are pure substances ? what are some examples ?}\\
            E & \textit{what are the characteristics of the elizabethan theater ?}\\
            \midrule
            H = 4 &  \textit{what are pure substances ?}\\
            H = 5 & \textit{what are some of pure substances ?}\\
            H = 6 & \textit{what are some examples of pure substances ?}\\
            H = 7 & \textit{what are some examples of a pure substance ?}\\
            \bottomrule
        \end{tabular}
        \caption{\label{tab:heightgen}Sample generations with different levels of syntactic control. S and E stand for source and exemplar, respectively. Please refer to \refsec{subsec:syngran} for details. }
    }
\end{table}

As can been seen in \reftbl{tab:heightgen}, at height 4 the syntax tree provided to the model is not enough to generate the full sentence that captures the meaning of the original sentence. As we increase the height to 5, it is able to capture the semantics better by predicting \textit{some of} in the sentence. We see that at heights 6 and 7 \methodmain{} is able to capture both semantics and syntax of the source and exemplar respectively.
 However, as we provide the complete height of the tree i.e., 7, it further tries to follow the syntactic input more closely leading to sacrifice in the overall relevance since the original sentence is about \textit{pure substances} and not \textit{a pure substance}.
\noindent It can be inferred from this example that since a source sentence and exemplar's syntax might not be fully compatible with each other, using the complete syntax tree can potentially lead to loss of relevance and grammaticality. Hence by choosing different levels of syntactic granularity, one can address the issue of compatibility to a certain extent. \newline

\noindent \textbf{2. Syntactic Variety : }\reftbl{tab:tempgen} shows sample generations of our model on multiple exemplars for a given source sentence. It can be observed that \methodmain{} can generate high-quality outputs for a variety of different template exemplars even the ones which differ a lot from the original sentence in terms of their syntax. A particularly interesting exemplar is \textit{what is chromosomal mutation ? what are some examples ?}. Here, \methodmain{} is able to generate a sentence with two question marks while preserving the essence of the source sentence. It should also be noted that the exemplars used in \reftbl{tab:tempgen}, were selected manually from the test sets, considering only their \textit{qualitative compatibility} with the source sentence. Unlike the procedure used for the creation of QQP-Pos dataset, the final \textit{paraphrases} were not kept in hand while selecting the \textit{exemplars}. In real-world settings, where a \textit{gold paraphrase} won't be present, these results are indicative of the qualitative efficacy of our method.

\subsection{\methodmain{}-R Analysis}
\label{subsec:posbleu}
ROUGE based selection from the candidates favour paraphrases which have higher n-gram overlap with their respective source sentences, hence may capture source's meaning better.  This hypothesis can be directly observed from the results in \reftbl{tab:synpar_res} and \reftbl{tab:human_eval} where we see higher values on automated semantic and human evaluation scores. While this helps in getting better semantic generations, it tends to result in higher TED values. One possible reason is that, when provided with the complete tree, fine-grained information is available to the model for decoding and it forces the generations to adhere to the syntactic structure. In contrast, at lower heights, the model is provided with lesser syntactic information but equivalent semantic information.

\subsection{Qualitative Analysis}
\label{subsec:qual_analysis}
\begin{table}[htpb]
    \scriptsize{\centering
    \label{tab:label}
    \begin{tabular}{m{3em}P{4.7em}P{7.7em}P{5.5em}}
        \toprule
        & Single-Pass & Syntactic Signal & Granularity \\
        \toprule
        \scpn{} & \xmark & Linearized Tree &  \cmark \\
        \midrule
        \cgen{} & \cmark & POS Tags (During training) & \xmark \\

        \midrule
        \methodmain{} & \cmark & Constituency Parse Tree & \cmark \\

        \bottomrule
    \end{tabular}
\caption{\label{tab:compare} Comparison of different syntactically controlled paraphrasing methods. Please refer to \refsec{subsec:qual_analysis} for details.}}
\end{table}
As can be seen from \reftbl{tab:compare}, \methodmain{} not only incorporates the best aspects of both the prior models, namely  \scpn{} and \cgen{}, but also utilizes the complete syntactic information obtained using the constituency-based parse trees of the exemplar.

From the generations in \reftbl{tab:generations}, it can be observed that our model is able to capture both, the semantics of the source text as well as the syntax of template. \scpn{}, evidently, can produce outputs with the template syntax, but it does so at the cost of semantics of the source sentence. This can also be verified from the results in \reftbl{tab:synpar_res} where \scpn{} performs poorly on PDS as compared to other models. In contrast \cgen{} and \methodmain{} retain much better semantic information, as is desirable. While generating sentences, \cgen{} often abruptly ends the sentence as in example 1 in \reftbl{tab:generations}, truncating the penultimate token with \textit{of}. The problem of abrupt ending due to insufficient syntactic input length was highlighted in \citet{chen2019controllable} and we observe similar trends. \methodmain{} on the other hand generates more relevant and grammatical sentences.

Based on empirical evidence, \methodmain{} alleviates this shortcoming, possibly due to dynamic syntactic control and decoding. This can be seen in e.g., 3 in \reftbl{tab:generations} where \cgen{} truncates the sentence abruptly (penultimate token =  \textit{directors}) but \methodmain{} is able to generate relevant sentence without compromising on grammaticality.

\subsection{Limitations and Future directions}
\label{subsec:discussion}

All natural language English sentences cannot necessarily be converted to any desirable syntax. We note that \methodmain{} does not take into account the compatibility of source sentence and template exemplars and can freely generate syntax conforming paraphrases. This at times, leads to imperfect paraphrase conversion and nonsensical sentences like example 6  in \reftbl{tab:tempgen} (\textit{is career useful in software ?}). Identifying compatible exemplars is an important but separate task in itself, which we defer to future work.

Another important aspect is that the task of paraphrase generation is inherently domain agnostic. It is easy for humans to adapt to new domains for paraphrasing. However, due to the nature of the formulation of the problem in NLP, all the baselines as well as our model(s), suffer from dataset bias and are not directly applicable to new domains. A prospective future direction can be to explore it from the lens of domain independence.

Analyzing the utility of controlled paraphrase generations for the task of data augmentation is another interesting possible direction.

\section{Conclusion}
\label{sec:conclusion}

In this paper, we proposed \methodmain{}, an end-to-end framework for the task of syntactically controlled paraphrase generation. \methodmain{} generates paraphrase of an input sentence while conforming to the syntax of an exemplar sentence provided along with the input. \methodmain{} comprises a GRU-based sentence encoder, a modified RNN based tree encoder, and a pointer-generator based novel decoder. In contrast to previous works that focus on a limited amount of syntactic control, our model can generate paraphrases at different levels of granularity of syntactic control without compromising on relevance. Through extensive evaluations on real-world datasets, we demonstrate \methodmain{}'s efficacy over state-of-the-art baselines.

\noindent We believe that the above approach can be useful for a variety of text generation tasks including syntactic exemplar-based abstractive summarization, text simplification and data-to-text generation.

\section*{Acknowledgement}
\label{sec:ack}
This research is supported in part by the Ministry of Human Resource Development (Government of India). We thank the action editor Asli Celikyilmaz, and the three anonymous reviewers for their helpful suggestions in preparing the manuscript. We also thank Chandrahas for his indispensable comments on drafts of this paper.

\bibliography{tacl2019}
\bibliographystyle{acl_natbib}

\end{document}